\documentclass[runningheads]{llncs}
\usepackage{graphicx}
\usepackage{subfigure}
\begin{document}
\title{Learn to Enhance the Negative Information in Convolutional Neural Network}
%
%

\author{Zhicheng Cai\and 
Chenglei Peng$^\star$
\and
Qiu Shen\thanks{Corresponding author: Chenglei Peng and Qiu Shen
\\This work was supported by National Natural Science Foundation of China under Grant 62071216 and Natural Science Foundation of Jiangsu Province of China via Grant BK20211149.}
}
%
%
\institute{School of Electronic Science and Engineering, Nanjing University \\
\email{caizc@smail.nju.edu.cn, \{pcl, shenqiu\}@nju.edu.cn}
}

\maketitle              
\begin{abstract}
This paper proposes a learnable nonlinear activation mechanism specifically for convolutional neural network (CNN) termed as LENI, which learns to enhance the negative information in CNNs. In sharp contrast to ReLU which cuts off the negative neurons and suffers from the issue of ``dying ReLU'', LENI enjoys the capacity to reconstruct the dead neurons and reduce the information loss. Compared to improved ReLUs, LENI introduces a learnable approach to process the negative phase information more properly. In this way, LENI can enhance the model representational capacity significantly while maintaining the original advantages of ReLU. As a generic activation mechanism, LENI possesses the property of portability and can be easily utilized in any CNN models through simply replacing the activation layers with LENI block. Extensive experiments validate that LENI can improve the performance of various baseline models on various benchmark datasets by a clear margin (up to 1.24\% higher top-1 accuracy on ImageNet-1k) with negligible extra parameters. Further experiments show that LENI can act as a channel compensation mechanism, offering competitive or even better performance but with fewer learned parameters than baseline models. In addition, LENI introduces the asymmetry to the model structure which contributes to the enhancement of representational capacity. Through visualization experiments, we validate that LENI can retain more information and learn more representations.
\keywords{CNN \and Nonlinear Activation \and Computer Vision.}
\end{abstract}

\begin{figure}[t]
\centering
\subfigure[LENI Mechanism]{
      \includegraphics[width=0.54\textwidth]{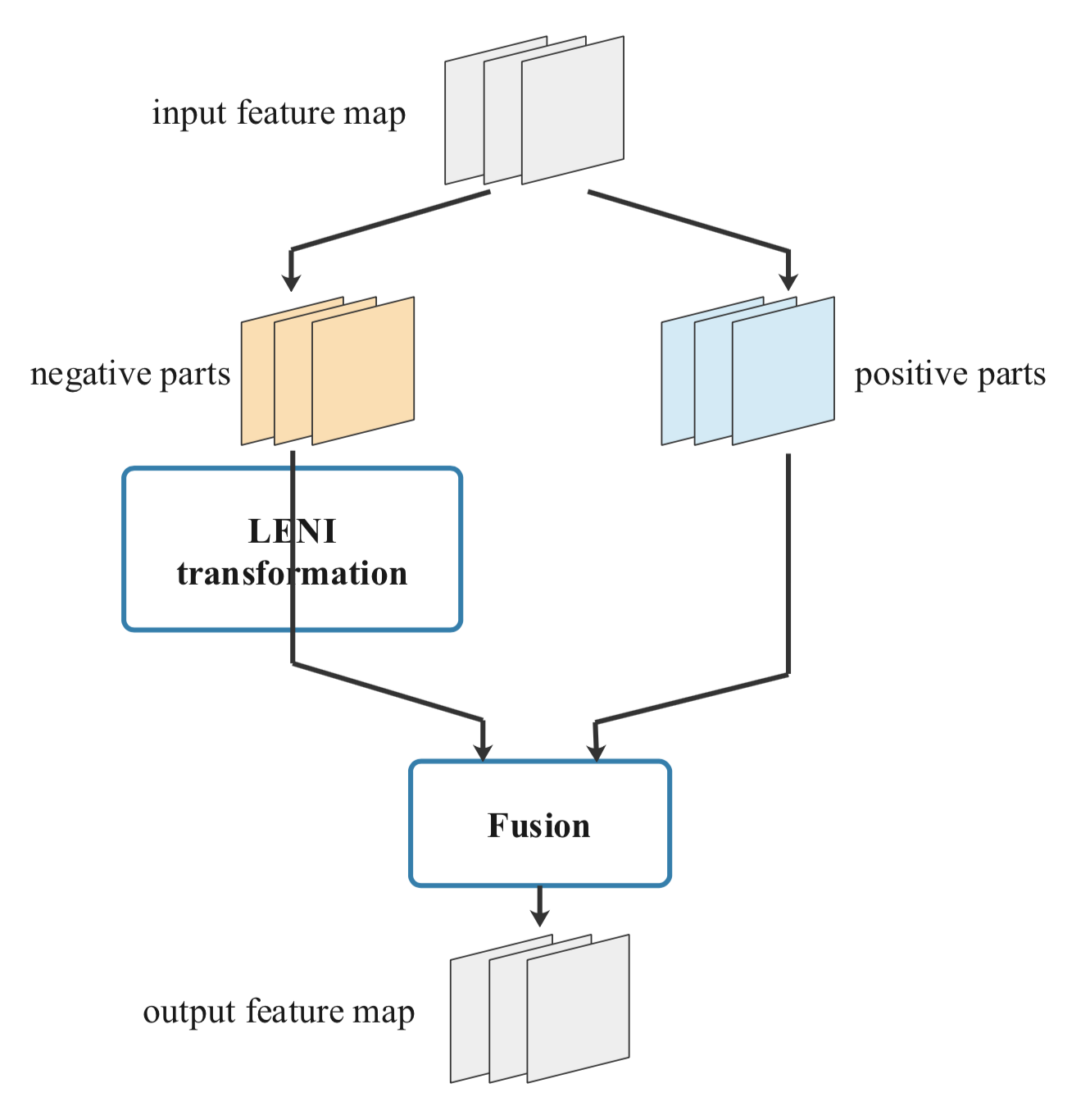}
      \label{fig1-1}
      }
\subfigure[LENI Block]{
      \includegraphics[width=0.40\textwidth]{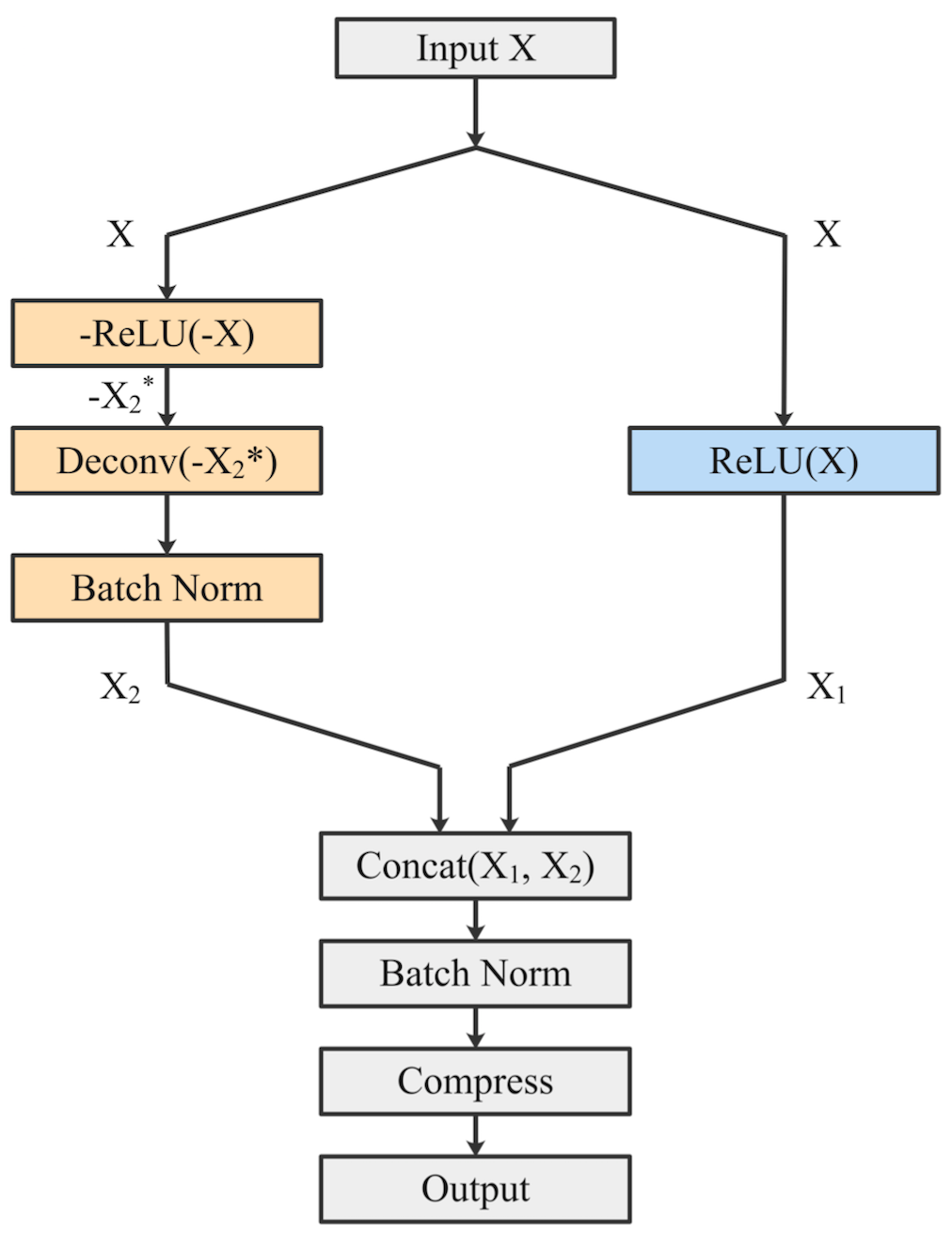}
      \label{fig1-2}
      }
\caption{LENI mechanism and LENI block}
\label{fig1}
\vspace{-1em}
\end{figure}

\section{Introduction}
Since AlexNet~\cite{krizhevsky2012imagenet} won the ILSVRC-2012, convolutional neural network (CNN) has flourished in a compelling way~\cite{szegedy2015going,he2016deep,howard2017mobilenets,sandler2018mobilenetv2,howard2019searching} and been successfully applied in many real scenarios.
It is known that the effective utilization of the information at all layers is vital for CNNs. Typically, CNN adopts ReLU~\cite{glorot2010deep} as the activation function. 
Although ReLU is an unsaturated nonlinear function and brings benefits for training dynamics, the forward propagation process will truncate all the negative neurons, thus the backward propagation will cause the death of these neurons. This issue is termed as ``dying ReLU''~\cite{Parisi_2022}.
Consequently, the information contained in these truncated neurons fails to be adopted effectively and the information with negative phase is completely obstructed through the network. Furthermore, the information loss will accumulate through layers, thus significantly weakening the model representational capacity.
While some improved ReLUs like Leaky ReLU~\cite{maas2013rectifier} take advantage of negative information, they inevitably lose some ReLU's original advantages. In addition, these functions may not process the negative phase information properly in that they simply multiply the negative values with relatively light weights. As proved in the later experiments, replacing ReLU with these improved ReLUs fails to improve the model's performance significantly.

This paper raises a novel nonlinear activation mechanism termed \textbf{LENI} specifically for CNN models. 
The core of LENI mechanism is utilizing an extra learnable transformation (termed as \textbf{LENI transformation}) to re-process the neurons cut off by ReLU. Subsequently, the information carried by the LENI neurons are fused with the positive feature maps to the further forward propagation through \textbf{Fusion}. Fig.~\ref{fig1-1} exhibits the pipeline of LENI mechanism. \textbf{LENI block} (as exhibited in Fig.~\ref{fig1-2}) is utilized to realize the LENI mechanism and replace the activation layers in CNN models. 
LENI aims to take better advantage of the information carried by all neurons and learn to process the negative phase information more properly, it reduces the loss of information in the network flow without sacrificing the benefits that ReLU provides. In this way, LENI can enhance the model representational capacity significantly. Extensive experiments validate that LENI can improve the performance of various models on various benchmarks datasets by a clear margin (e.g., up to 1.24\% higher top-1 accuracy on ImageNet-1k) with negligible extra parameters. Through visualization experiments, we validate that LENI can indeed retain more information and learn more representations.

Except for the effective utilization of information and the enhancement of representational capacity, LENI possesses other advantages. 
First, LENI block can act as a channel compensation mechanism~\cite{shang2016understanding}, thus it can enhance the utilization effectiveness of convolution kernels and achieve the same performance with fewer convolution kernels (namely, fewer parameters). 
Second, the asymmetrical structure of the LENI block can break the symmetry of CNN models, which increases the rank of the weight matrices and further improves the model representational capacity. 
Third, LENI block has a nice portability and can be easily implemented in various CNN baseline models on the mainstream deep learning frameworks like PyTorch. 
In addition, we argue that LENI is more than a nonlinear activation function typified by ReLU, to be specific, it is a nonlinear activation mechanism which introduces learnable parameters, possesses certain structure, and conducts the nonlinear activation operation. 
\section{Related Work}
\subsection{Nonlinear Activation Functions}
The nonlinear activation function follows a weight layer to conduct nonlinear operation, allowing neural networks to learn the nonlinear mapping. A common and effective nonlinear activation function is the Rectified Linear Unit (ReLU)~\cite{glorot2010deep}.
The gradient of ReLU is either one for positive neurons or zero for negative neurons, which alleviates the issue of vanishing gradient. Besides, ReLU can accelerate the convergence and make the optimization more stable\cite{Parisi_2022}.
Although ReLU brings many benefits, it discards the negative neurons and executes simple truncation to the negative information, while negative neurons account for a large proportion~\cite{Parisi_2022} and the negative information can possess valuable features~\cite{Parisi_2022}, thus ReLU results in the loss of information and weakens the model representational capacity. This issue is termed as ``dying ReLU''~\cite{Parisi_2022}.
In an attempt to mitigate the ‘dying ReLU’ issue, some researchers improve ReLU by taking advantage of negative values, such as LeakyReLU~\cite{maas2013rectifier}, PReLU~\cite{he2015delving} and so on~\cite{XuWCL15,Barron17a,clevert2015fast,hendrycks2016gaussian}. 
However, these improved ReLUs simply multiply the negative values with light weights, which is explored insufficiently and may not be a proper treatment for the negative phase information. 
Besides, empirical investigations~\cite{cai2021study} validate that under certain circumstance, improved ReLUs fail to perform better than ReLU, and they inevitably lose some original advantages of ReLU.

\subsection{Utilization of the Negative Information}
Because of ReLU, the negative values are abandoned and the information flow in CNN is constantly positive. However, this unbalanced circumstance maybe improper for that the negative information can be valuable. Kuoet al.~\cite{8456277,JAYKUO2018237,KUO2019346} pointed that ReLU makes CNN have its preference on images over their foreground and background reversed ones, which is termed as the issue of \emph{sign confusion}. To tackle this issue, Saak~\cite{8456277} augments the kernels by duplicating the negative ones. Shang et al.~\cite{shang2016understanding} proved that at the shallow layers of CNN, it is sensitive to both positive and negative information, tending to capture these two kinds of information simultaneously. Thus CReLU~\cite{shang2016understanding} concatenated both positive and negative information simply along the channels. Although LENI may be slightly similar to CReLU, there are two great differences. Firstly, CReLU changes the width of the original model while LENI can keep the model width unchanged. Secondly, CReLU simply channel-concatenates the positive and negative feature maps without any learnable process, while LENI introduces extra trainable transformation on the negative feature maps and intends to enhance the negative information and process the negative information properly. 


\section{Method}
\subsection{LENI Mechanism and LENI Block}
To process the negative neurons more properly and enhance the valuable information in the negative phase, we rethink the flow mode of negative phase information in deep CNN models and introduce a novel nonlinear activation mechanism termed LENI.

The module that introduces and implements the LENI mechanism is termed as LENI block, as Fig.~\ref{fig1} exhibits. 
Suppose $X$ is the input of LENI block. LENI block has two parallel branches. One branch is a single ReLU, it selects the positive neurons of $X$ and truncates the negative ones as a common practice, outputting $X_1$ with positive information.
Simultaneously, $-X$ is input into another paralleled ReLU to screen out the negative neurons and obtains $X_2^*$  with negative information. In order to retain the phase of the gradient, $X_2^*$ is inverted. Then a trainable transformation is conduct on $-X_2^*$ to process the negative information and obtain $X_2$. In LENI block, the certain transformation consists of a depthwise separable deconvolution layer and a batch normalization layer. 
After that, $X_1$ and $X_2$ are concatenated along the channels and then batch normalized. To compress the doubled channels, one $1\times1$ convolution layer is employed after the fusion.
Finally, the nonlinearly activated feature map is obtained. 

In this way, negative neurons otherwise perished can be re-connected and re-processed through certain trainable transformation, therefore the negative phase information is able to propagate forward and backward effectively.
LENI is superior to a function which can be simply written in the formation of normal function expression and drawn the function curve. Consequently, this operation is considered as a kind of nonlinear activation mechanism.

For implementation, we can simply replace all the original activation functions with LENI without fine-tuning the feature channels or other hyper-parameters. Moreover, we observed that only utilizing the LENI in the shallow layers (or only the first layer) and keeping the rest activation layers unchanged can still enhance the accuracy higher up to $1.24\%$ for MobileNetV3~\cite{howard2019searching} on ImageNet dataset.

\subsection{Advantages of LENI}
\textbf{Maintain the advantages of ReLU.} LENI has two parallel branches. One of the branch only contains a single ReLU, retaining ReLU's advantages undoubtedly. The other branch conducts the ReLU before deconvolution, which works in the same way as the ReLU before the next convolution layer in normal CNN, thus still retaining ReLU's advantages.
Besides, the concatenation and compression will not damage ReLU's advantages. As a result, LENI retains the ReLU's advantages. 

\textbf{Effective usage of information and enhancement of representational capacity.} It has been validated that features with negative phase possesses valuable information as well~\cite{maas2013rectifier,shang2016understanding}. Instead of simply discarding the negative phase information like ReLU, or multiplying negative values with slight weights like improved ReLUs, or simply concatenating negative and positive values like CReLU, LENI conducts trainable transformation on the negative feature maps and fuses them with the positive feature maps. Thus LENI offers a more proper way to process and exploit the valuable negative phase information, which enables the CNN model to take full advantage of the features and reduce the loss of information, thus enhancing the model representational capacity. 

\textbf{Asymmetrical structure design.} LENI block has an asymmetrical structure, breaking the model symmetrical characteristic, which is beneficial to enhancing the model representational capacity. Suppose that only a small number of neurons alter their activated values according to the different inputs while the majority of the neurons are insensitive to the different inputs, thus the order of the weight matrix is small, and the total order will be significantly decayed after the multiplication of weight matrices. Consequently, although the dimension of weight matrices is high, most of the dimensions possess little valuable information, thus weakening the model representational capacity. Such degradation problem is partially attributed to the symmetrical characteristic of the model. 

\textbf{Channel compensation mechanism.} LENI is a channel compensation mechanism that enhances the efficiency of feature channels. As stated above, the lower layers of CNN are sensitive to both positive and negative information~\cite{shang2016understanding}, while ReLU forbids the negative phase information to flow, thus requiring convolution kernels with more redundancy channels. This issue is termed as \emph{complementary phenomenon of parameters}~\cite{shang2016understanding}. LENI alleviates this problem by processing the negative phase information with certain transformation and fusing both positive and negative information. Consequently, the model with LENI can utilize the kernels more effectively and achieve the same performance as that of the model with ReLU with a fewer number of parameters, which will be verified in the later experiments. 

\textbf{Portability.} LENI is a generic activation mechanism and can be utilized in any CNNs. For implementation, we can simply utilize the LENI block to replace the original activation layers without fine-tuning the feature channels or other hyper-parameters. In addition, LENI can be easily realized on the mainstream deep learning frameworks like PyTorch. 
\begin{table}[t]
\caption{Test accuracy of various methods on ImageNet. The number of parameters and inference speeds of these methods are also exhibited. The speed is tested on 4$\times$3090 Ti GPUs with a batch-size of 256, measured in examples/second.}
\centering
\begin{tabular}{|l|l|c|c|c|c|c|c|}
\hline
Model \ \ \ \ \ & \ \ \ LENI  & \ Top1-Acc \ & \ \ \ \ $\uparrow$\ \ \ \  &  \ \ Params\ \  & \ \ \ \ $\uparrow$\ \ \ \  & \ \ Speed\ \ & \ \ \ \ $\downarrow$\ \ \ \   \\
\hline
VGGNet-16     & \ \ \ \ \ \ -- --  & 73.34\% & -- --  & 15.236M    & -- --  & 1383/s &-- -- \\
VGGNet-16     & layer 1            & 73.82\% &+0.48\% & 15.246M    &+0.010M & 1375/s &\ \ $-$8/s\\
VGGNet-16     & layer 1,2          & 73.85\% &+0.51\% & 15.256M    &+0.020M & 1362/s &$-$21/s \\
VGGNet-16     & layer 1,2,3        & 73.92\% &+0.58\% & 15.266M    &+0.030M & 1347/s &$-$36/s \\
\hline
ResNet-18     & \ \ \ \ \ \ -- --  & 70.94\% & -- --  & 11.724M    & -- --  & 1492/s &-- -- \\
ResNet-18     & layer 1            & 71.29\% &+0.35\% & 11.734M    &+0.010M & 1479/s & $-$13/s \\
ResNet-18     & layer 1,2          & 71.25\% &+0.31\% & 11.744M    &+0.020M & 1468/s & $-$24/s \\
ResNet-18     & layer 1,2,3        & 71.35\% &+0.41\% & 11.754M    &+0.030M & 1457/s & $-$35/s \\
\hline
MobileNetV1   & \ \ \ \ \ \ -- --  & 72.08\% & -- --  & \ \ 4.232M & -- --  & 1583/s & -- --\\
MobileNetV1   & layer 1            & 72.44\% &+0.36\% & \ \ 4.234M &+0.002M & 1571/s &$-$12/s \\
MobileNetV2   & \ \ \ \ \ \ -- --  & 71.68\% & -- --  & \ \ 3.565M & -- --  & 1753/s &-- --\\
MobileNetV2   & layer 1            & 71.99\% &+0.31\% & \ \ 3.567M &+0.002M & 1737/s &$-$16/s\\
MobileNetV3-S & \ \ \ \ \ \ -- --  & 61.95\% & -- --  & \ \ 2.938M & -- --  & 1578/s &-- --\\
MobileNetV3-S & layer 1            & 63.19\% &+1.24\% & \ \ 2.939M &+0.001M & 1573/s &\ \ $-$5/s \\
MobileNetV3-L & \ \ \ \ \ \ -- --  & 71.61\% & -- --  & \ \ 5.476M & -- --  & 1723/s &-- --\\
MobileNetV3-L & layer 1            & 71.98\% &+0.37\% & \ \ 5.477M &+0.001M & 1721/s &\ \ $-$2/s \\
\hline
\end{tabular}
\label{t1}
\end{table}

\begin{table}[t]
\caption{Test accuracy of three baseline CNNs with different activation methods on various benchmark datasets.} 
\centering
\begin{tabular}{|l|c|c|c|c|c|c|c|}
\hline
Method & CIFAR-10 & CIFAR-100 & \ \ SVHN\ \  & MNIST & KMNIST & FMNIST & Params \\
\hline
\multicolumn{8}{|c|}{VGGNet-11}\\
\hline
ReLU & 90.39\%& 67.27\% & 93.96\%& 99.50\% &97.06\%  &92.95\% &5.21M\\
\hline
Leaky ReLU & 90.43\% & 66.89\% & 93.76\% & 99.42\% &96.64\% &93.09\%&5.21M\\
PReLU & 89.38\%&66.30\% & 93.64\%& 99.52\% &96.97\% &92.84\%&5.21M\\
RReLU & 90.66\%& 68.09\% & 94.05\%& 99.41\% &97.09\% &93.20\%&5.21M\\
CELU & 89.45\%& 65.62\% & 93.57\%& 99.41\% &96.59\% &92.76\%&5.21M\\
GELU & 87.92\%& 61.13\% & 93.50\%& 99.41\% &96.25\%&92.65\%&5.21M\\
ELU & 90.00\% & 65.92\% & 93.42\%& 99.41\% &96.24\%&92.82\%&5.21M\\
CReLU & 89.83\% & 66.71\%&  94.37\% & 99.41\% & 97.68\% & 93.02\% &9.39M\\
\hline
LENI & \textbf{91.80\%} & \textbf{68.41\%} & \textbf{95.34\%} & \textbf{99.59\%} &\textbf{98.24\%} &\textbf{93.66\%} &6.02M\\ 
\hline
\multicolumn{8}{|c|}{VGGNet-16}\\
\hline
ReLU & 91.58\%& 68.48\% & 94.82\%& 99.55\% &97.55\%  &93.36\%&15.24M\\
\hline
Leaky ReLU & 91.52\% & 68.89\% & 94.86\% & 99.51\% &97.47\% &93.52\%&15.24M\\
PReLU & 91.08\%&68.11\% & 94.55\%& 99.50\% &97.55\% &93.71\%&15.24M\\
RReLU & 91.75\%& 69.38\% & 94.74\%& 99.48\% &97.69\% &93.48\%&15.24M\\
CELU & 89.55\%& 68.67\% & 94.57\%& 99.47\% &97.43\% &92.79\%&15.24M\\
GELU & 87.61\%& 65.02\% & 94.27\%& 99.41\% &97.20\%&92.71\%&15.24M\\
ELU & 90.22\% & 66.34\% & 94.31\%& 99.43\% &97.39\%&92.92\%&15.24M\\
CReLU & 92.06\% & 69.07\%&  95.37\% & 99.61\% &97.78\% & 93.39\%&29.68M\\
\hline
LENI & \textbf{93.17\%} & \textbf{70.77\%} & \textbf{95.81\%} & \textbf{99.70\%} &\textbf{98.51\%} &\textbf{93.86\%}&21.52M\\ 
\hline
\multicolumn{8}{|c|}{ResNet-34}\\
\hline
ReLU & 93.64\%& 74.89\% & 95.43\%& 99.57\% &98.17\%  &94.23\%&21.33M\\
\hline
Leaky ReLU & 93.56\% & 74.37\% & 95.41\% & 99.51\% &98.17\% &94.17\%&21.33M\\
PReLU & 93.46\%&72.68\% & 95.45\%& 99.50\% &97.95\% &94.33\%&21.33M\\
RReLU & 93.77\%& 74.91\% & 95.53\%& 99.48\% &98.09\% &94.35\%&21.33M\\
CELU & 92.52\%& 72.34\% & 95.47\%& 99.47\% &97.63\% &93.76\%&21.33M\\
GELU & 90.64\%& 70.22\% & 94.93\%& 99.41\% &97.40\%&93.88\%&21.33M\\
ELU & 92.71\% & 72.86\% & 95.01\%& 99.43\% &97.49\%&93.91\%&21.33M\\
CReLU & 93.89\% & 75.32\%&  95.67\% & 99.61\% &98.34\% & 94.17\%&41.85M\\
\hline
LENI & \textbf{94.73\%} & \textbf{76.56\%} & \textbf{96.72\%} & \textbf{99.73\%} &\textbf{99.05\%} &\textbf{95.01\%}&28.44M\\
\hline
\end{tabular}
\label{t2}
\end{table}
\section{Experiment}
\subsection{Performance Improvements on ImageNet}
To verify the effectiveness of LENI, we compare the performance of various baseline models with LENI or ReLU on ImageNet. ImageNet is the most challenging and authoritative benchmark dataset for image classification, which comprises $1.28$M images for training and $50$K for validation from $1000$ classes. The baselines include VGGNet-16~\cite{simonyan2014very}, ResNet-18~\cite{he2016deep}, MobileNetV1~\cite{howard2017mobilenets}, MobileNetV2~\cite{sandler2018mobilenetv2}, and MobileNetV3~\cite{howard2019searching} with ReLU. Consider that the shallow layers of CNN are more sensitive to both positive and negative information, we only employ LENI to replace the ReLUs in shallow layers here, which also considers the trade off between the performance gain and extra computational costs.
The training configuration follows the benchmark~\cite{he2016deep}.

Table~\ref{t1} shows the experimental results. As can be observed, the performance of all models with LENI is consistently improved by a clear margin.
For example, when only replacing ReLU with LENI in the single first activation layer, the Top-1 accuracy of MobileNetV3-S is lifted significantly by $1.24\%$. When applying LENI in the first three activation layers, VGGNet-16 and ResNet-18 obtain the accuracy improvement of $0.58\%$ and $0.41\%$ respectively. 
The results validate that LENI can enhance the model representational capacity. In addition, LENI is cost-efficient considering that the improvement is significant while the amount of the extra parameters is so slight that can be ignored. 
\subsection{Comparison with Other Activation Functions}
We further conduct comparative experiments of baseline CNNs with different nonlinear activation methods. We simply replace all the ReLUs in baseline models with LENI or other improved activation functions including CReLU, Leaky ReLU, PReLU, RReLU~\cite{XuWCL15}, CELU~\cite{Barron17a}, GELU~\cite{hendrycks2016gaussian}, and ELU~\cite{clevert2015fast}. The baseline models include VGGNet-11, VGGNet-16 and ResNet-34. After that, we employed the same benchmark training configurations as \cite{he2016deep} to train these CNN models on six benchmark datasets, namely, CIFAR-10, CIFAR-100, SVHN, MNIST, KMNIST, and Fashion-MNIST. These datasets are described detailedly in \cite{cai2021jitter}.

As Table~\ref{t2} shows, LENI enabled the CNN models to obtain significantly better results than all the other activation functions. Moreover, we can see that some improved ReLU functions performed slightly better or even poorer than the original ReLU function. The experimental results verify that LENI can enhance the model representational capacity and process the negative phase information in a more proper way. 
\begin{table}[t]
\caption{Test accuracy of three baseline CNNs, of which the width is halved, with ReLU, CReLU and LENI, and test accuracy of baselines with extended convolution layers.}
\centering
\begin{tabular}{|l|c|c|c|c|c|c|r|}
\hline
Method & CIFAR-10 & CIFAR-100 & \ \ SVHN \ \  & MNIST & KMNIST & FMNIST & Params \\
\hline
\multicolumn{8}{|c|}{VGGNet-11}\\
\hline
ReLU & 90.39\%& 67.27\% & \textbf{93.96\%}& 99.50\% &97.06\%  &92.95\% & 5.21M\\
\hline
Half ReLU & 88.91\%& 65.22\% & 92.91\%& 99.43\% &96.11\%  &92.45\% &1.30M\\
Half CReLU & 88.62\% &65.01\%&  93.56\% & 99.36\% & \textbf{97.15\%} & 92.93\% &2.35M\\
Half LENI & \textbf{91.19\%}& \textbf{67.35\%} & 93.90\%& \textbf{99.53\%} &96.92\%  &\textbf{93.05\%} &1.51M\\
\hline
Extended Conv & 90.45\%& 66.79\% & 93.93\%& 99.54\% &96.92\%  &93.17\% &6.02M\\
\hline
\multicolumn{8}{|c|}{VGGNet-16}\\
\hline
ReLU & 91.58\%& 68.48\% & 94.82\%& 99.55\% &97.55\%  &93.36\% & 15.24M\\
\hline
Half ReLU & 90.45\%& 67.25\% & 94.67\%& 99.42\% &97.56\%  &92.56\% &3.81M\\
Half CReLU & 90.56\%& 67.10\% & 95.11\%& 99.51\% &97.61\%  &92.74\% &7.42M\\
Half LENI & \textbf{91.74\%}& \textbf{68.56\%} & \textbf{95.56\%}& \textbf{99.63\%} &\textbf{98.27\%}  &\textbf{93.39\%} &5.37M\\
\hline
Extended Conv & 90.54\%& 67.32\% & 94.96\%& 99.54\% &97.75\%  &93.09\% &21.52M\\
\hline
\multicolumn{8}{|c|}{ResNet-34}\\
\hline
ReLU & \textbf{93.64\%}& 74.89\% & 95.43\%& 99.57\% &98.17\%  &\textbf{94.23\%} &21.33M\\
\hline
Half ReLU & 92.65\%& 72.66\% & 94.23\%& 99.56\% &97.87\%  &93.57\% &5.33M\\
Half CReLU & 92.76\% & 73.14\%&  95.21\% & 99.54\% &98.04\% & 93.91\% &10.46M\\
Half LENI & 93.15\%& \textbf{75.23\%} & \textbf{95.73\%}& \textbf{99.62\%} &\textbf{98.22\%}  &93.97\% &7.10M\\
\hline
Extended Conv & 93.67\%& 74.93\% & 95.36\%& 99.57\% &98.21\%  &94.25\% &28.44M\\
\hline
\end{tabular}
\vspace{-1em}
\label{t3}
\end{table}
\subsection{Channel Compensation Mechanism}
We further design experiments that halve the model width to validate the channel compensation of LENI. 
Table~\ref{t3} shows the experimental results, where mark \textbf{Half} means the number of feature channels in all convolution layers of the model is halved. 
It is observed that although the model width is halved, the models with LENI still performs better than the models with ReLU which possess three times more parameters, demonstrating that LENI can enhance the utilization efficiency of the feature channels.
Compared to the CReLU which is also tested as another channel compensation method, LENI obtains better results. This is attributed to that LENI conducts well-designed operation on the negative channels thus processes the negative feature channels more properly, while CReLU simply concatenates negative channels with the positive ones without any processing operations.
\subsection{Further Analysis about Effectiveness}
It is noted that LENI block contains more learned parameters. To validate that the more learned parameters is not the key reason for the effectiveness of LENI, we add one $3\times3$ depthwise convolution layer immediately after each ReLU in baselines.
Row \textbf{Extended Conv} in Table~\ref{t3} shows the experimental results. It is observed that these extended convolution layers fail to improve the representational capacity of the baseline models. This experimentally illustrates that the structural asymmetry that LENI block introduces is functional.
In addition, CReLU widens the model by two times and doubles the parameter amount as exhibited in Table~\ref{t2}. CReLU utilizes both positive and negative information and is asymmetric. However, it still fails to improve the model performance significantly. This experimentally illustrates that the deconvolutional learnable transformation that LENI block introduces is vital.
Thus it can be concluded that more parameters do not necessarily improve the model representational capacity which makes no contribution to the effectiveness of LENI. Consequently, we attribute the effectiveness of LENI to the proper processing (realized by certain trainable transformation) of the negative information and the well-designed asymmetric structure of LENI block.
\begin{figure}[!t]
  \centering
  \subfigure[Non-activated]{
      \includegraphics[width=0.42\textwidth]{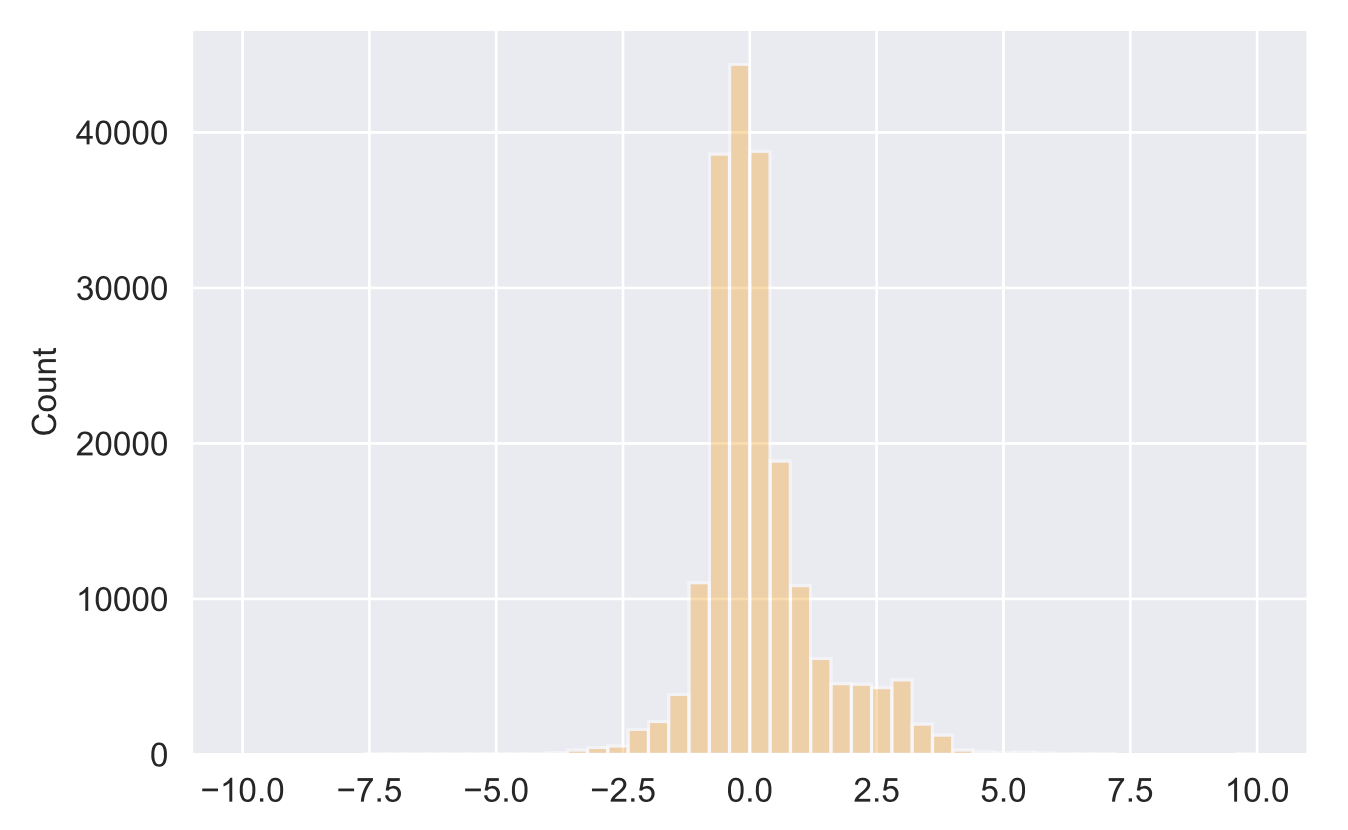}
      \label{dis-1}
      }
  \subfigure[ReLU-activated]{
      \includegraphics[width=0.42\textwidth]{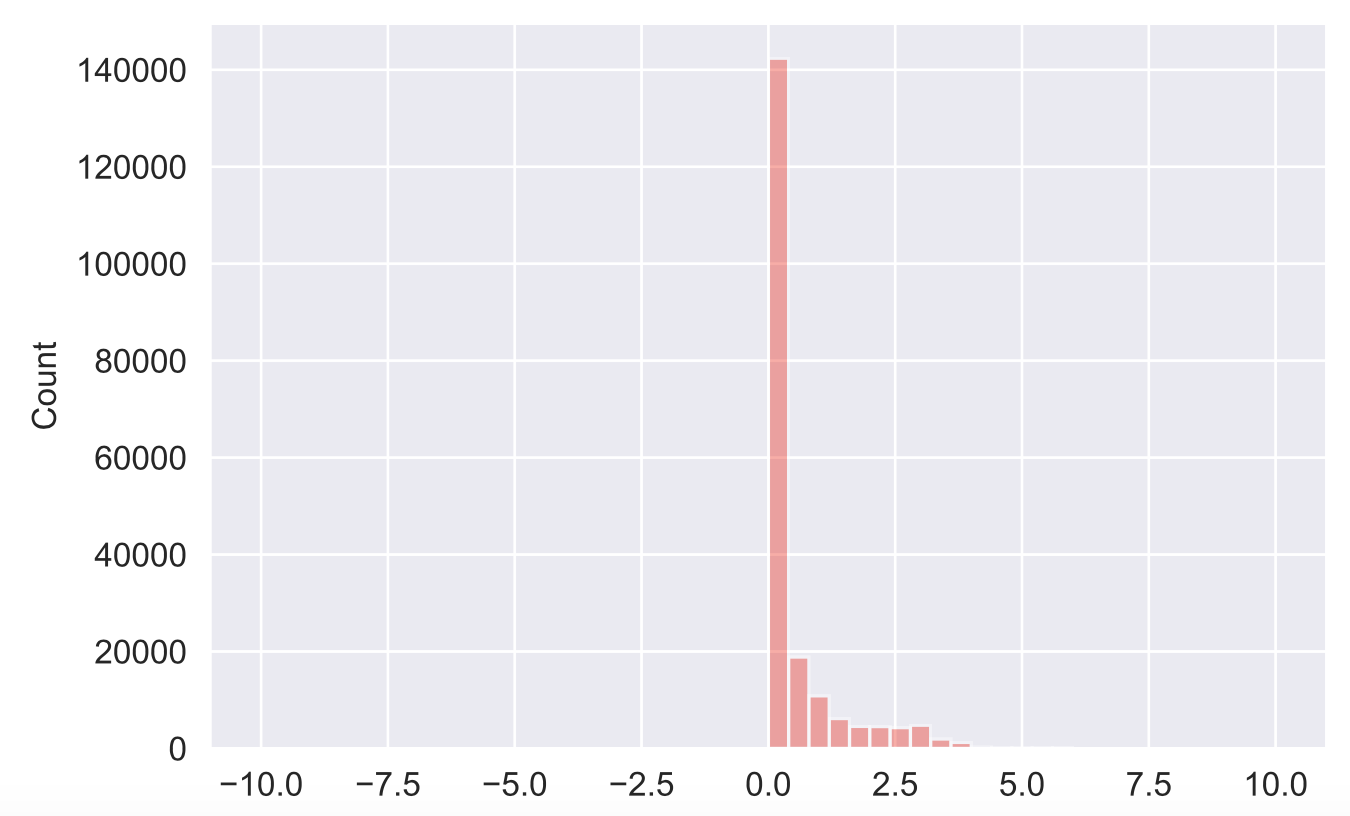}
      \label{dis-2}
      }
  \subfigure[Leaky ReLU-activated]{
      \includegraphics[width=0.42\textwidth]{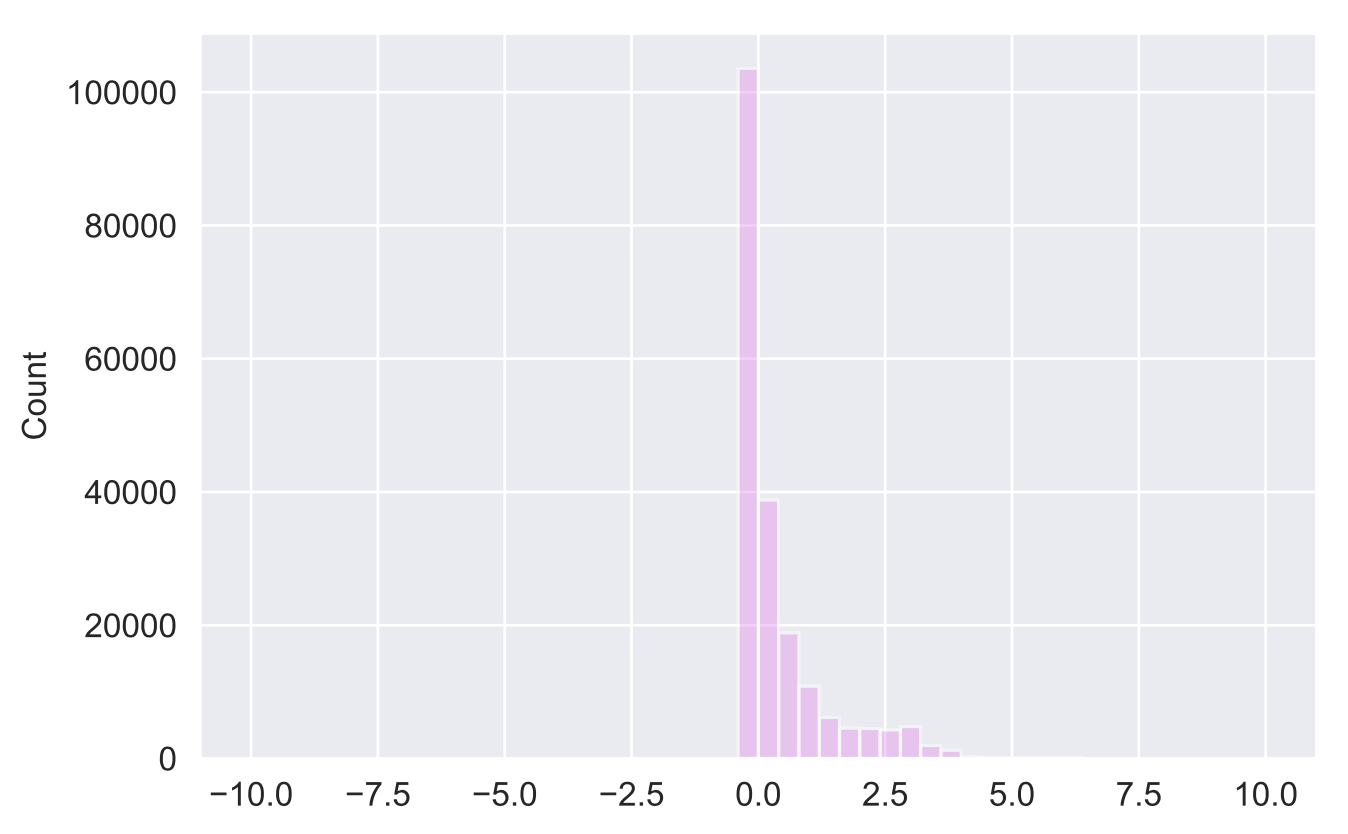}
      \label{dis-3}
      }
  \subfigure[LENI-activated]{
      \includegraphics[width=0.42\textwidth]{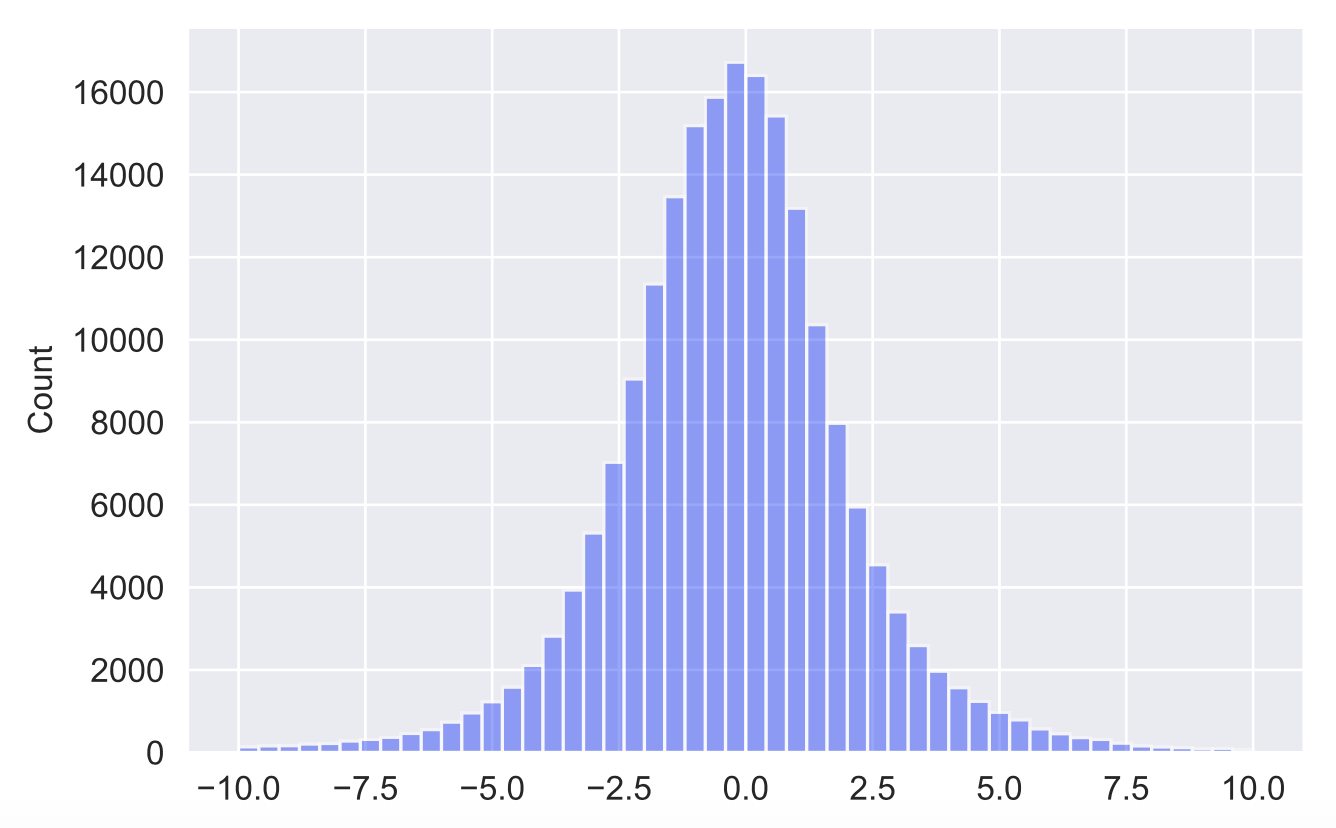}
      \label{dis-4}
      }
  \caption{Distribution histogram of the values in the feature maps non-activated or activated by ReLU, Leaky ReLU and LENI. About half of the neurons are truncated by ReLU, which loses much information. For Leaky ReLU, the negative values are multiplied with a small coefficient of 0.01, which makes the negative values close to zero, thus the distribution histogram is similar to that of ReLU. For LENI, it is similar to the normal distribution, which is the distribution possessing the most information entropy, thus validating that LENI can retain more information. }
  \label{dis}
\vspace{-1em}
\end{figure}
\subsection{LENI Retains More Information}\label{sec-dis}
Fig.~\ref{dis} exhibits the distribution histogram of the values in the feature maps non-activated or activated by ReLU, Leaky ReLU and LENI. 
The baseline model utilized is the MobileNetV3-S trained on ImageNet.
We extract the pre-activated and activated feature maps of the first convolution layer. 
It can be observed that about half of the values in the non-activated feature maps are negative and truncated by ReLU, leading to the loss of much information. 
For Leaky ReLU, these negative values are multiplied with a small coefficient of 0.01, which makes the negative values cluster close to zero, thus the distribution histogram is observed similarly to that of ReLU. 
However, for LENI, the distribution histogram is meaningful as it is similar to the normal distribution, which is the distribution possessing the most information entropy. This phenomenon validates that LENI can retain more information, which can be attributed to the extra learnable transformation that LENI conducts to enhance the negative information.
\begin{figure}[t]
   \centering
   \includegraphics[width=0.98\textwidth]{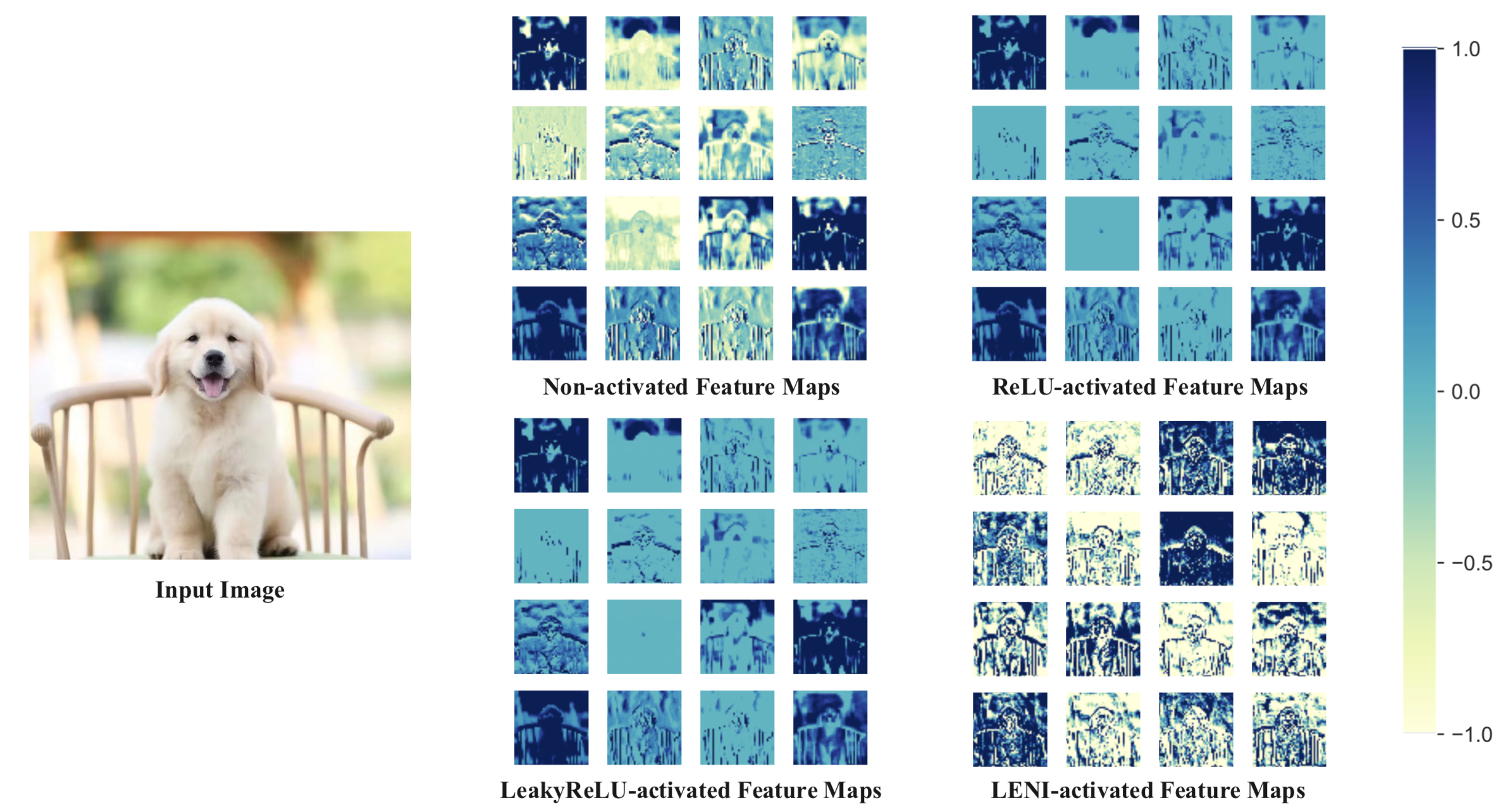}
   \caption{Visualization of the feature maps non-activated or activated by ReLU, Leaky ReLU and LENI. The pixel with darker color possesses larger value.  It is obviously that, compared to the non-activated feature maps, the feature maps activated by ReLU lose many representations of significance due to the ``dying ReLU" issue. Leaky ReLU still has the similar issue that losing meaningful representations. While for LENI, it is observed that the representations are more richer, for example, the edges are more clear and the textures are more distinct. This phenomenon illustrates that LENI can learn more representations through the additional learnable transformation that conducts the proper processing on the negative information.}
   \label{pic}
\vspace{-1em}
\end{figure}
\subsection{LENI Learns More Representations}\label{vis}
Fig.~\ref{pic} visualizes various non-activated feature maps and the feature maps activated by ReLU, Leaky ReLU and LENI. Still, the baseline model utilized is the MobileNetV3-S trained on ImageNet and we extract the pre-activated and activated feature maps of the first convolution layer. 
It is obviously that, compared to the non-activated feature maps, the feature maps activated by ReLU lose many representations of significance due to the ``dying ReLU" issue, specifically, the contrast of object and background is blurry and many patterns are discarded.
The feature maps activated by Leaky ReLU is similar to that of ReLU (the reason is the same as stated above), there still exists the similar issue that losing meaningful representations. Consequently, Leaky ReLU fails to obtain better representational capacity than ReLU.
While for LENI, it can clearly observed that the representations are more richer, for example, the edges are more clear and the textures are more distinct.
This phenomenon illustrates that LENI can learn more representations through the additional learnable transformation that conducts the proper processing on the negative information.
\section{Conclusion}
This paper proposes LENI, an innovative nonlinear activation mechanism, to learn to enhance the negative information and strengthen the model representational capacity. As a generic design element, LENI block can be implemented in various CNN baselines. Extensive experiments validate that LENI can produce superior performance to ReLU or improved ReLUs. Further experiments illustrate that LENI can enhance the efficiency of convolution kernels. We attribute the effectiveness of LENI to the proper processing of negative information and the well-designed asymmetric structure. Through visualization experiments, we validate that LENI can retain more information and learn more representations.

{\small

}
\end{document}